\documentclass[conference]{IEEEtran}
\IEEEoverridecommandlockouts
\usepackage{cite}
\usepackage{amsmath,amssymb,amsfonts}
\usepackage{algorithmic}
\usepackage{graphicx}
\usepackage{textcomp}
\usepackage{xcolor}
\usepackage{url}
\usepackage{algorithm}
\usepackage{orcidlink}
\usepackage{subcaption}
\def\BibTeX{{\rm B\kern-.05em{\sc i\kern-.025em b}\kern-.08em
    T\kern-.1667em\lower.7ex\hbox{E}\kern-.125emX}}

\begin{document}

\title{Integrating Background Knowledge in Medical Semantic Segmentation with Logic Tensor Networks}

\author{\IEEEauthorblockN{Luca Bergamin}
\IEEEauthorblockA{\textit{Department of Mathematics} \\
\textit{University of Padua}\\
Padova, Italy \\
bergamin[at]math.unipd.it \orcidlink{0000-0002-0662-7862} }
\and
\IEEEauthorblockN{Giovanna Maria Dimitri}
\IEEEauthorblockA{\textit{Department of Information Engineering and Mathematics} \\
\textit{University of Siena}\\
Siena, Italy \\
 giovanna.dimitri[at]unisi.it \orcidlink{0000-0002-2728-4272}}
\and
\IEEEauthorblockN{Fabio Aiolli}
\IEEEauthorblockA{\textit{Department of Mathematics} \\
\textit{University of Padua}\\
Padova, Italy \\
fabio.aiolli[at]math.unipd.it \orcidlink{0000-0002-5823-7540} }
}

\maketitle

\begin{abstract}
Semantic segmentation is a fundamental task in medical image analysis, aiding medical decision-making by helping radiologists distinguish objects in an image. Research in this field has been driven by deep learning applications, which have the potential to scale these systems even in the presence of noise and artifacts. However, these systems are not yet perfected. We argue that performance can be improved by incorporating common medical knowledge into the segmentation model’s loss function. To this end, we introduce Logic Tensor Networks (LTNs) to encode medical background knowledge using first-order logic (FOL) rules. The encoded rules span from constraints on the shape of the produced segmentation, to relationships between different segmented areas. We apply LTNs in an end-to-end framework with a SwinUNETR for semantic segmentation. We evaluate our method on the task of segmenting the hippocampus in brain MRI scans. Our experiments show that LTNs improve the baseline segmentation performance, especially when training data is scarce. Despite being in its preliminary stages, we argue that neurosymbolic methods are general enough to be adapted and applied to other medical semantic segmentation tasks.

\end{abstract}


\section{Introduction}
\label{sec:introduction}

Semantic segmentation is a classic task in medical image analysis.
{The task involves classifying each pixel in an image into a predefined category \cite{long2015fully}. Unlike object detection, which identifies and localizes objects with bounding boxes, semantic segmentation provides a detailed understanding of scene composition by assigning meaningful labels to every pixel. This fine-grained image analysis is crucial for various applications, including medical imaging \cite{ronneberger2015u,InfoFusion}, autonomous driving \cite{chen2017deeplab}, satellite imagery analysis, and environmental monitoring.}
{For what concerns medical imaging, several are the possible areas of} applications, such as tumor segmentation, organ segmentation, and lesion segmentation.
{One of the main issues that affect semantic segmentation is the dimensionality of the data used for training. It is, in fact, true that it would usually require quite a large amount of labeled data to perform well. This represents many times, a particularly challenging point in medical imaging, where the availability of large amounts of data is sometimes challenging \cite{medicalSegmentation}.
This is, in fact, due to the high cost of annotation and the privacy concerns that are proper of the medical data fields.}
This is the reason why developing a semantic segmentation method that can also leverage background knowledge to improve the segmentation performance is fundamental. In this field, neuro-symbolic AI has been emerging \cite{3rdwave} as a promising avenue to create new AI systems that are human-centric \cite{tango}, arguing how the synergy of machines and people is key to achieving better results for critical systems, such as in healthcare.

\begin{figure}[tp]
    \centering
    \begin{minipage}{0.45\columnwidth}
        \centering
        \includegraphics[width=\textwidth]{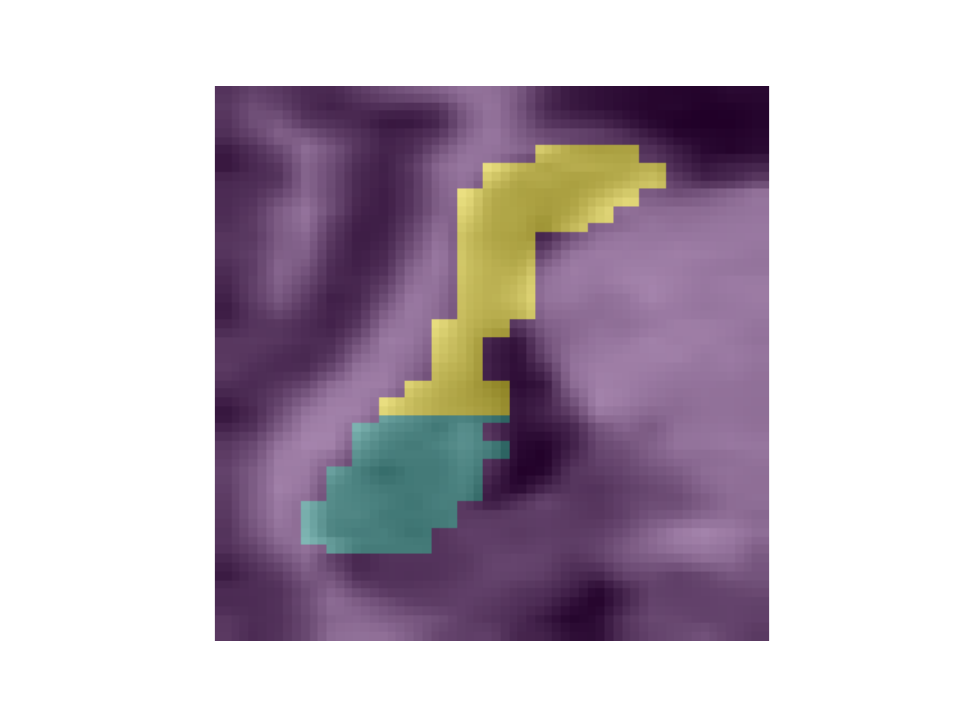}
        \subcaption{Ground truth data. The yellow and green areas are clearly distinct.}
        \label{fig:fig1}
    \end{minipage}  \hfill
    \begin{minipage}{0.45\columnwidth}
        \centering
        \includegraphics[width=\textwidth]{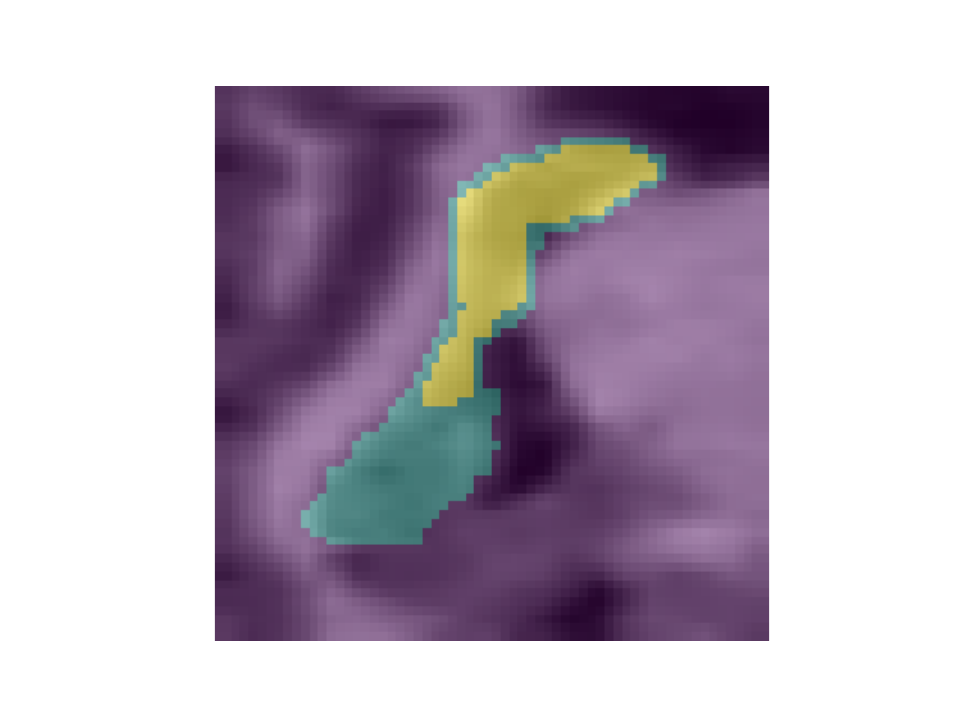}
        \subcaption{Predicted data. The green border around the yellow area is implausible.}
        \label{fig:fig2}
    \end{minipage}
    \caption{Motivating example for this work.}
    \label{fig:main}
    \vspace{-1.5em}

\end{figure}

In this paper, we propose a proof of concept of semantic segmentation for the brain hippocampus, using 3D MRI images and adding to semantic segmentation deep AI networks additional logical rules, which can help significantly in providing further information for completing the segmentation task. More specifically, we made use of the well-known  SwinUNETR segmentation network, enriched with the Logic Tensor Network (LTN) environment, to add logical rules within the learning process of the semantic segmentation network \cite{LTN}.

LTN is a powerful tool for modeling background knowledge in the form of propositional logic.
In particular, LTN can define soft logical constraints integrated into a loss function that can be used to guide the learning process of a neural network.
In this paper, we propose a novel method to integrate background knowledge in medical semantic segmentation with LTN. To the best of our knowledge, no such work has ever been proposed before for hippocampus segmentation. As mentioned, we evaluated our method on the task of segmenting the hippocampus in brain MRI scans as a proof of concept, leveraging the background knowledge that has been gathered through common medical knowledge and manual inspection of the outputs of the model.

The rest of the paper is organized as follows.
In Section~\ref{sec:related_work}, we review the related work in both the fields of semantic segmentation in brain imaging and LTN. In Section~\ref{sec:method}, we present our method with the relevant context for the baseline models and frameworks used. In Section~\ref{sec:experiments}, we describe the experiments performed. In Section~\ref{sec:results}, we present the results.
Finally, in Section~\ref{sec:conclusion}, we conclude the paper.

\section{Related Work}
\label{sec:related_work}

\subsection{Brain Hippocampus Segmentation Background}
\label{sec:BrainHippoSegmentation}

{Artificial intelligence (AI) has significantly advanced the field of medical imaging, particularly for tasks such as the segmentation of anatomical structure \cite{MedicalSegmentation}. 
More specifically, in this field, particular relevance has been observed in the context of brain structuring and anatomy segmentation. In this paper, we focus on the task of automatically segmenting the region of the hippocampus. This represents, in fact, a critical brain structure involved in memory and learning \cite{Hippo}. Accurate segmentation of the hippocampus region is essential for diagnosing and monitoring neurological conditions such as Alzheimer's disease \cite{HippoAlzheimer}. Several AI-based methodologies have been developed to enhance the precision and efficiency of hippocampal segmentation. Most AI deep learning-based approaches rely on the use of convolutional neural network-based architectures. For instance, in \cite{DeepHipp} the authors employ 3D CNNs integrated with attention mechanisms. More specifically, the DeepHipp model utilizes a 3D dense-block architecture combined with attention mechanisms to effectively extract relevant features from MRI scans, achieving high accuracy in hippocampal segmentation \cite{DeepHipp}. Generative adversarial networks (GANs) have also been applied to this task. By leveraging the adversarial training process, GANs can, in fact, generate more accurate segmentation maps, capturing subtle details of hippocampal subfields. An example of the application of such an approach is \cite{GAN_Hippocampus}, a method that has shown improved performance over traditional CNN-based techniques. Other alternative strategies rely instead on the use of multi-model deep learning frameworks that combine segmentation and classification tasks. For example in \cite{MultiModel_CNN} the authors use a multi-task CNN can perform hippocampal segmentation while simultaneously classifying Alzheimer's disease status, thereby enhancing the overall diagnostic process. Additionally, the U-Net architecture \cite{UNet}, known for its efficacy in biomedical image segmentation, has been adapted for hippocampal segmentation. Modifications to the original U-Net, such as altering kernel sizes, have led to improved segmentation performance, achieving an average accuracy of 96.5\% as works as for instance \cite{UNet_Hippocampus} can witness. Furthermore, integrating high-resolution T2-weighted MRI data into deep learning models has been shown to enhance segmentation accuracy. By using T2-informed deep CNNs, researchers have achieved superior performance compared to traditional methods, facilitating more reliable assessments of hippocampal atrophy \cite{T2Informed_CNN, HIPPO2, HIPPO3, HIPPO4}.

\subsection{Background knowledge integration in semantic segmentation}

While a large number of methods have been proposed for semantic segmentation, few of them have considered integrating background knowledge with a logic language.
Among the few that have, Knowledge-Enhanced Neural Networks (KENN, \cite{kenn}) is a notable example applied to semantic cloud segmentation, albeit not applied in medical applications, but to 3d natural scenes. The authors consider the integration of many hand-engineered feature extractors, and they include some expert knowledge relevant to the task at hand (e.g., a rooftop is on top of a house). The authors observe that including this external knowledge helps boost the model performance, especially in cases where the class of interest has few examples and/or takes few pixels in the image. Our results confirm this hypothesis, showing also how model performance is impacted across different data set sizes. 
For what concerns medical imaging segmentation only a few examples exist. For instance in \cite{NeuroSymbolicMedicalImaging} the authors present an approach where deep neural networks for semantic segmentation are enhanced through a neurosymbolic approach in the context of intracranial aneurysm. However no approaches, to the best of our knowledge, for the segmentation of the hippocampus have been proposed.

\section{Method}
\label{sec:method}

In this section, we present our method to integrate background knowledge in medical semantic segmentation with LTN. Firstly, we discuss the semantic segmentation task. Secondly, we consider Swin Transformers. Finally, we introduce Logic Tensor Networks, and our integration of the aforementioned topics.

\subsection{Semantic Segmentation Task}
\label{subsec:segmentation}

Generally speaking, semantic segmentation is a task that aims to assign a class label to each pixel in an image.  More specifically, by assigning a categorical label to each pixel in an image, we get a partition of the image into semantically meaningful regions. Formally, given an input image \( I: \Omega \to \mathbb{R}^c \), where \( \Omega \subset \mathbb{Z}^2 \) denotes the set of pixel coordinates and \( c \) represents the number of color channels, the goal is to learn a mapping function \( f: \mathbb{R}^c \to \{1, \dots, K\} \) that maps each pixel to one of \( K \) predefined semantic classes. The resulting segmentation map \( S: \Omega \to \{1, \dots, K\} \) which assigns a unique class label to each pixel, where \( S(p) = k \) indicates that pixel \( p \in \Omega \) belongs to class \( k \).  

Deep learning-based approaches to semantic segmentation typically employ an encoder-decoder architecture, where the encoder extracts hierarchical feature representations from the input image, and the decoder performs spatial upsampling to generate a dense prediction map.

\subsection{Swin Transformers and Swin-UNETR}
\label{subsec:swinunetr}

{Among the latest proposed architectures for performing semantic segmentation, Swin-UNETR is one of the state-of-the-art works in this field \cite{SwinUNET}, which we also employed in our work. Swin-UNETR is a variant of the SwinTransformer \cite{SwinTransformer} that has been adapted for semantic segmentation. The SwinTransformer is an advanced vision transformer model designed for image processing tasks. Unlike traditional vision transformers (ViTs) that operate on full-sized image patches, Swin Transformers introduce a hierarchical structure that enables computational efficiency and better feature extraction. It employs shifted windows for self-attention, reducing computational complexity while maintaining global context awareness \cite{SwinTransformer}. Swin-UNET was optimized explicitly for medical image segmentation. More specifically, Swin-UNET incorporates the hierarchical self-attention mechanism of Swin Transformers within a U-Net-like encoder-decoder framework to enhance medical imaging tasks \cite{SwinUNET}. Swin-UNET operates by first dividing an input image into non-overlapping patches. Such patches are then embedded into feature representations using a linear embedding layer. The architecture consists of multiple Swin Transformer blocks that process hierarchical representations while maintaining local and global feature dependencies. The self-attention in a Swin-UNET can be expressed as:
\begin{equation}
    \text{Attention}(Q, K, V) = \text{softmax}\left(\frac{QK^T}{\sqrt{d_k}}\right) V
\end{equation}
where $Q$, $K$, and $V$ represent query, key, and value matrices, respectively, and $d_k$ is the dimensionality of the keys. The shifted window mechanism allows for an overlapping receptive field across layers, ensuring better feature fusion across different spatial locations. This approach enhances segmentation accuracy while maintaining efficiency. Concerning the decoder in Swin-UNET, it follows a symmetric path, where extracted features are upsampled using transposed convolutions and concatenated with encoder features via skip connections. The final segmentation map is then generated using a convolutional layer with a softmax activation function:
\begin{equation}
    P(y | x) = \frac{e^{f_i(x)}}{\sum_j e^{f_j(x)}}
\end{equation}
where $P(y | x)$ represents the probability of a pixel belonging to class $y$, given input $x$.

\subsection{Logic Tensor Network}
\label{subsec:ltn}

Logic Tensor Network (LTN) is a framework that combines the flexibility of neural networks with a sound method to evaluate and enforce codified background knowledge \cite{LTN}.

LTN leverages a specific logic language named Real Logic \cite{LTN2}, which is fully differentiable and can be seamlessly integrated into deep learning architectures.

To compute real values that can be used during backpropagation, a mapping function $\mathcal{G}$ is used. This function, named \textit{grounding function}, makes it possible to manipulate logic expressions in a neural network in a sound framework.

For example, a common way to implement the $\forall$ operation is by using a smooth minimum function:

\begin{equation}
    \mathcal{G} (\forall) : u_1, ..., u_n \mapsto 1- \left(\frac{1}{n} \sum_{i=0}^n (1-u_i)^p\right)^{1/p}
\end{equation}

Once an LTN grounding is defined, one can express a knowledge base KB $\mathcal{K}=\{\phi_1, ..., \phi_k\}$ using logical connectives and quantifiers. Finally, the model can be optimized by minimizing the following aggregated satisfaction loss, which corresponds to assuming that all the formulas in the knowledge base should be verified.

\begin{equation}
   \mathcal{L}(\theta) = 1 - \text{SatAgg}_{\phi \in \mathcal{K}} \mathcal{G}(\phi | \theta)
\end{equation}

\subsection{Integration of LTN and SwinUNETR}
\label{subsec:integration}

We integrate the LTN and SwinUNETR by defining soft logical constraints that guide the learning process of the SwinUNETR.

We consider the following logical constraints:

\begin{itemize}
    \item The anterior portion is connected to the posterior portion.
    \item The anterior portion cannot contain the posterior portion.
    \item The posterior portion cannot contain the anterior portion.
    \item The volume of the anterior and posterior portions are similar.
\end{itemize}

In the usual LTN fashion, first, we define the predicates used in the logical constraints.

\begin{itemize}
\item $\text{Dice($x$, $y$)}$: the value of the loss function of Dice for the output.
    \item $\text{Connected}(\hat{y})$: the predicted segmented regions $\hat{y}$ have the anterior and posterior portion connected.
    \item $\text{Nested}(\hat{y})$: the predicted segmented regions $\hat{y}$ are contained within another.
    \item $\text{SimVol}(x,y, \epsilon)$: $x$ has a similar volume to $y$.
\end{itemize}

The actual implementation of these functions follows:

\paragraph{Dice loss} The main target of the optimization aims to maximize the Dice score, i.e., how well the predicted pixels align to the ground truth data. This metric is explained in detail in the experimental section of this paper. To optimize this value, the following, differentiable approximation of the original Dice score is used:

\begin{equation}
    \mathcal{L}_{\text{dice}}(p,t) = 1- \frac{2 \sum p_i t_i}{\sum p_i \sum t_i},
\end{equation}

where $p_i$ and $t_i$ are the predicted value for the i-th pixel, and its corresponding target value, respectively. Note that the actual loss subtracts the approximated Dice score from 1, to allow its maximization with gradient descent.

\paragraph{Connection constraint} We compute the distance of the two segmentation masks via the Chamfer distance between two sets of points \( A \) and \( B \). This is defined as:

\[
d_{\text{Chamfer}}(A, B) = \frac{1}{|A|} \sum_{a \in A} \min_{b \in B} \| a - b \|^p + \frac{1}{|B|} \sum_{b \in B} \min_{a \in A} \| b - a \|^p,
\]

\begin{equation}
    \mathcal{G}(\phi_c; \gamma_c) = \exp \left( - \gamma_c \cdot (d(A,B))^2\right),
\end{equation}

we refer to this value as \textbf{connectedness} in the experimental section.

\paragraph{Nesting Constraint} We approximate the nesting constraint by sampling $P$ pairs of points within the same segmented region. For each pair, we verify whether any of the $Q$ points in the linear interpolant region still belong to the same class, as described in Algorithm~\ref{alg:nesting}. We refer to this value as \textbf{nesting} in the experimental section.

\begin{algorithm}
    \begin{algorithmic}
\STATE \textbf{Input:} 
\STATE \hspace{0.3cm} $A$, $B$: masks
\STATE \hspace{0.3cm} $P$, \text{the number of samples to draw}
\STATE \hspace{0.3cm} $Q$, \text{granularity value for interpolation}
\STATE \textbf{Output:} 
\STATE \hspace{0.5cm} \text{True} if, in the linear path from two points in $A$, there is a point in $B$, false otherwise. 

\STATE \text{Find indices where } \texttt{A == 1.}
\IF{number of such indices is less than 2} 
    \STATE \text{return False}
\ENDIF
\STATE \text{Randomly sample $P$ pairs of points (src, dst).}
\FOR{each sampled pair \((\text{src}, \text{dst})\)}
    \STATE \text{Extract spatial coordinates of src and dst, denoted as } \text{src\_coords} \text{ and } \text{dst\_coords.}
    \STATE \text{Linearly interpolate between src and dst, get $Q$ values.}
    \STATE \text{\textbf{for} each interpolated point:}
    \STATE \hspace{0.4cm} \text{Clamp the rounded coordinates to a valid range.}
    \STATE \hspace{0.4cm} \text{Check if the point corresponds to a 1 in \texttt{B}.}
    \IF{the point in \texttt{B} equals 1}
        \STATE \text{return True}
    \ENDIF
\ENDFOR

\STATE \text{return False}
\end{algorithmic}
\caption{Nesting estimation. }
\label{alg:nesting}

\end{algorithm}

\paragraph{Volume constraint} We codify that the volume $v(A)$ and $v(B)$ should be close, within a tolerance value $\epsilon$, where $c(x)$ is the number of nonzero pixels for a given mask $x$:

\begin{equation}
    \mathcal{G}(\phi_v; \gamma_v) = \exp \left( -\gamma_v \cdot (\max\left(\mid \text{c}(x) - \text{c}(y)\mid - \epsilon, 0 \right)^2 \right),
\end{equation}

we refer to this value as \textbf{volume similarity} in the experimental section. Finally, we integrate the LTN and SwinUNETR by adding the following soft logical constraints.

\begin{align}
    \phi_d &= \forall x, y \in \mathcal{X}, \mathcal{Y} &\quad 1-\text{DiceLoss}(x,y) \\
    \phi_c &= \forall \hat{y} \in {f(\mathcal{X}}) &\quad \text{MinDst}(\hat{y}) = 0 \\
    \phi_v &= \forall \hat{y} \in {f(\mathcal{X}}) &\quad \text{SimVol}(\hat{y}) \\
    \phi_n &= \forall \hat{y} \in f(\mathcal{X}) &\quad 1 - \text{Nested}(\hat{y})
\end{align}

Where $\mathcal{X}$ and $\mathcal{Y}$ are the batched input images and ground truth labels, respectively, and $f$ is the function codified by the Swin-UNETR model. Finally, the knowledge base is compiled in the usual way as follows.

\begin{align}
\mathcal{K}=\{\phi_d, \phi_c, \phi_v, \phi_n\},\\
   \mathcal{L}(\theta) = 1 - \text{SatAgg}_{\phi \in \mathcal{K}} \mathcal{G}(\phi | \theta)
\end{align}

\section{Experiments}
\label{sec:experiments}

In this section, we describe the experiments we conducted to evaluate our method. First, we discuss technical details for the dataset; next, we consider the Dice coefficient calculation. Finally, we provide necessary information to understand and potentially replicate our results.

\subsection{Dataset}

We used the Hippocampus dataset, gathered from the Medical Decathlon dataset challenge \cite{decathlon}, which consists of brain MRI scans with manual annotations of the hippocampus. Three classes are available:

\begin{itemize}
    \item Background (0) – Everything that is not part of the hippocampus.
\item Hippocampus Anterior (1) – The anterior (head) part of the hippocampus.
\item Hippocampus Posterior (2) – The posterior (body and tail) part of the hippocampus.

\end{itemize}

The dataset contains $394$ samples, where each sample is a 3D MRI scan of the brain, with a resolution of $64\times 64 \times 64$. The scans were acquired using a 3T MRI scanner with a voxel size of $1 \times 1 \times 1$ mm. The manual annotations were performed by expert radiologists.

\subsection{Evaluation Metric}

The performance of the segmentation model is assessed using the Dice coefficient \cite{dice1945measures}, a widely used metric for evaluating the overlap between a predicted segmentation mask and the ground truth. The Dice coefficient for a given class \( c \) is defined as:

\begin{equation}
    \text{Dice}_c = \frac{2 |\hat{Y}_c \cap Y_c|}{|\hat{Y}_c| + |Y_c|}
\end{equation}

where \( Y_c \) denotes the ground truth binary mask for class \( c \), and \( \hat{Y}_c \) represents the corresponding predicted segmentation mask. The numerator quantifies the number of correctly segmented pixels (true positives), while the denominator accounts for the total number of pixels in both the predicted and ground truth masks.

In the case of pixel-wise representations, where segmentation masks are encoded as binary matrices, the Dice coefficient can be expressed as:

\begin{equation}
    \text{Dice}_c = \frac{2 \sum \hat{Y}_c Y_c}{\sum \hat{Y}_c + \sum Y_c}
\end{equation}

For multi-class segmentation, the Dice coefficient is computed for each class independently and averaged to obtain the final score:

\begin{equation}
    \text{Dice} = \frac{1}{C} \sum_{c=1}^{C} \text{Dice}_c
\end{equation}

where \( C \) is the total number of classes.

The Dice coefficient ranges from 0 to 1, where a value of 1 indicates perfect agreement between the prediction and the ground truth, while a value of 0 signifies no overlap. Unlike conventional accuracy metrics, Dice is particularly suitable for medical image segmentation as it mitigates the impact of class imbalance and provides a robust measure of spatial overlap.

\begin{figure}
    \centering
    \includegraphics[width=0.7\linewidth]{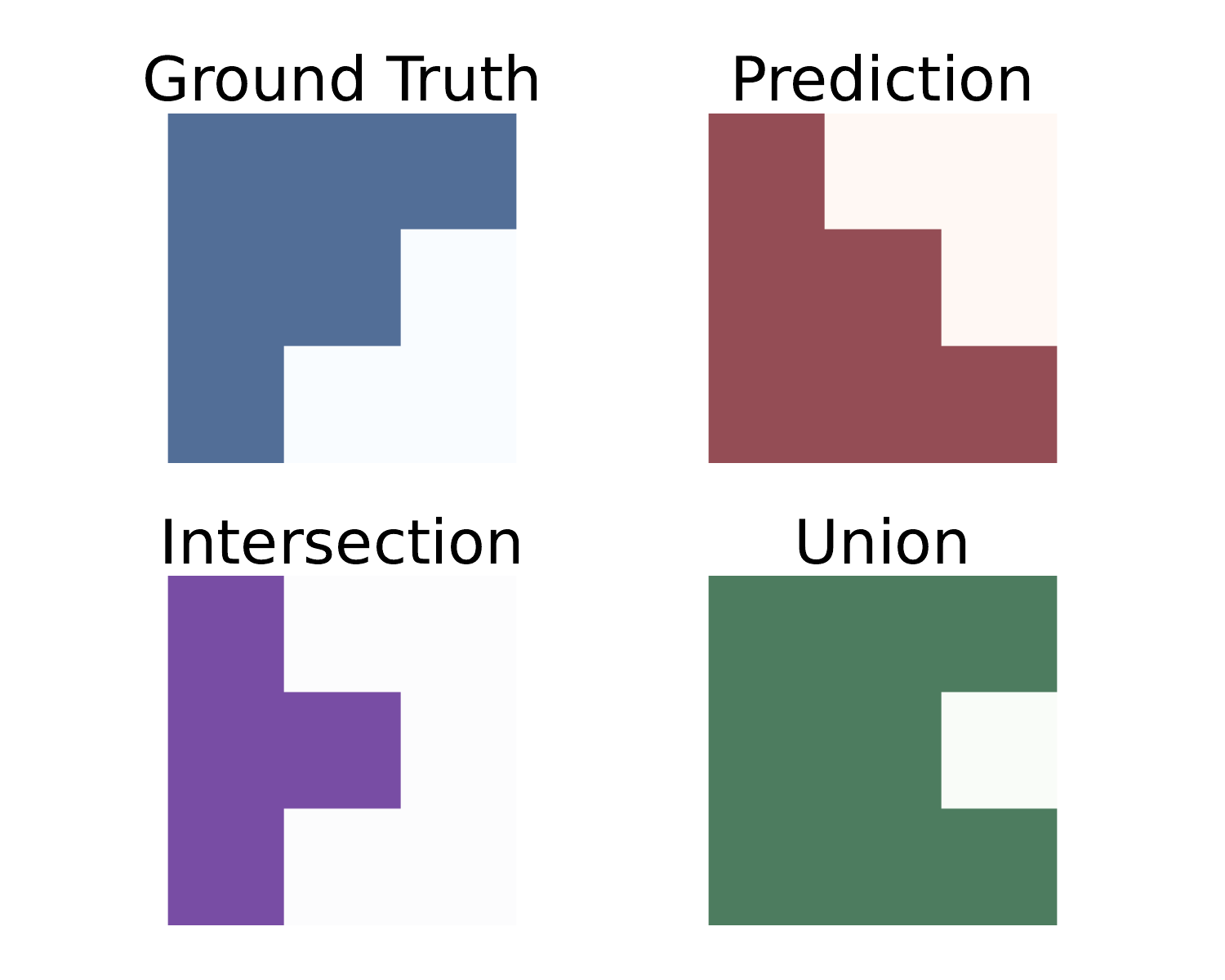}
    \caption{Dice coefficient example calculation. Here, Dice = $\frac{2 \cdot 4}{6+6} \simeq 0.666$.}
    \label{fig:enter-label}
\end{figure}

\subsection{Implementation and experimental setup}

We used publicly available implementations for LTN \cite{LTN2}\footnote{\url{github.com/tommasocarraro/ltntorch}}, and the MonAI framework\footnote{\url{github.com/Project-MONAI/MONAI}} to process the dataset and build the baseline SwinUNet model. As hyperparameters, we use a learning rate $\eta=0.0001$, with a warmup cosine annealing schedule, with a batch size $b=4$. For our methods, we fix $\gamma_v=0.0001$, $\epsilon=5000$, and $\gamma_c=0.001$. 

We cross-validate our results using $k=5$, thus splitting our data in an 80-20 proportion each time. Moreover, we consider different training dataset sizes, ranging from $1.0$ (full dataset after the train-val split), to $0.25$, and $0.05$, which correspond to $25\%$ and $5\%$ of the dataset, respectively.

We train each model for $E=100$ epochs. The baseline uses default hyperparameters, such as $\text{feature\_size}=24$, $\text{num\_heads}=(3,6,12,24)$, and $\text{depth}=(2,2,2,2)$.  No data augmentation method was employed. We make our Pytorch-based code implementation available on Github. \footnote{\url{github.com/BouncyButton/segmentation-with-ltn}}

\section{Results and Discussion}
\label{sec:results}

In this section, we present and analyze the results of our experiments, focusing on segmentation performance and constraint satisfaction. We provide both quantitative evaluations and qualitative insights to assess the impact of Logic Tensor Networks (LTN) in the context of medical image segmentation.

\subsection{Segmentation Performance}

\begin{table}[t]
\centering
\caption{Segmentation performance for the baseline and LTN models. We report in brackets the portion of the training set used.}
\label{tab:model_performance}
\scalebox{0.91}{
\begin{tabular}{|c|c|c|c|}
\hline
\textbf{Model} & \textbf{Dice (1.0)} & \textbf{Dice (0.25)} & \textbf{Dice (0.05)} \\
\hline
SwinUNet & 0.8547 ± 0.0127 &  0.8331 $\pm$ 0.0402 & 0.7434 ± 0.1123  \\
SwinUNet + LTN & \textbf{0.8628 ± 0.0157}  &  \textbf{0.8391 $\pm$ 0.0180}   & \textbf{0.7721 ± 0.0316}\\
\hline
\end{tabular}
}
\end{table}

Table~\ref{tab:model_performance} reports the segmentation performance of the baseline and LTN-based models, measured using the Dice coefficient. Across all dataset sizes, we observe a consistent improvement when incorporating LTN. Notably, the advantage is most pronounced in data-scarce settings, where deep learning models traditionally struggle due to limited training examples. In these cases, the explicit encoding of background knowledge within LTN serves as an additional learning signal, effectively guiding the optimization process towards semantically meaningful segmentations.

When using only 5\% of the training data, the baseline SwinUNet model exhibits a significant drop in performance, with a Dice score of $0.7434 (\pm 0.1123)$. However, the introduction of LTN mitigates this decline, yielding a Dice score of $0.7721 (\pm 0.0316)$, an improvement of approximately 3 percentage points. This suggests that LTN acts as a form of regularization, preventing the model from overfitting to the limited training samples and enabling better generalization. The performance gain remains observable at 25\% of the dataset size, though it diminishes as the dataset size increases. When training with the full dataset, the SwinUNet model achieves a Dice score of $0.8547 (\pm 0.0127)$, while the LTN-augmented version reaches $0.8628 (\pm 0.0157)$, still demonstrating a measurable, albeit smaller, benefit.

These results highlight an important characteristic of LTN: while its most significant impact is observed in low-data regimes, it also provides a small yet consistent boost in performance even when sufficient training data is available. This suggests that logical constraints not only compensate for data scarcity but also contribute to learning representations that better align with domain-specific knowledge. Importantly, these improvements are achieved without requiring additional labeled examples, making LTN particularly valuable in medical imaging tasks where annotated data is often expensive and time-consuming to acquire.

\subsection{Constraint Satisfaction}

While the Dice coefficient provides a useful global measure of segmentation quality, it does not capture the structural plausibility of the predicted segmentations. In medical imaging, even small deviations from anatomical correctness can lead to clinically significant errors. For example, as illustrated in Fig.~\ref{fig:fig2}, segmentations that contain disconnected regions or violate known spatial relationships may not be reflected in standard accuracy metrics but are nonetheless problematic from a medical standpoint. 

To further investigate the impact of LTN, we evaluate the satisfaction of key structural constraints. Specifically, we analyze: (i) \textbf{connectedness}, which measures whether the two segmented regions are attached; (ii) \textbf{nesting violations}, which quantify the occurrence of anatomically implausible  relationships, where one region inglobes another; and (iii) \textbf{volume similarity}, which assesses whether the two predicted segmentation maintains realistic proportions.

\begin{table}[t]
\centering
\caption{Constraint satisfaction for the baseline and LTN models. We report the model trained with the full dataset.}
\label{tab:constraint_satisfaction}
\begin{tabular}{|c|c|c|c|}
\hline
\textbf{Model} & \textbf{Connected ($\uparrow$)} & \textbf{Nested ($\downarrow$)} & \textbf{SimVol ($\uparrow$)} \\
\hline
Ground Truth data & 0.9741 $\pm$ 0.0002 & 0.0721 $\pm$ 0.0478 & {1.0 $\pm$ 0.0} \\
\hline
SwinUNet & 0.9744 ± 0.0004 &  0.4322 $\pm$ 0.0985 & \textbf{1.0 $\pm$ 0.0} \\
SwinUnet + LTN & \textbf{0.9746 $\pm$ 0.0008}  &  \textbf{0.3357 $\pm$ 0.1607} & \textbf{1.0 $\pm$ 0.0} \\
\hline
\end{tabular}
\end{table}

Table~\ref{tab:constraint_satisfaction} presents the results for these metrics, evaluated on models trained with the full dataset. The results indicate that the LTN-augmented model improves constraint satisfaction across all metrics. The most notable improvement is observed in the nesting violation metric, which decreases from $0.4322 (\pm 0.0985)$ for the baseline model to $0.3357 (\pm 0.1607)$ with LTN. This suggests that logical constraints successfully encourage more anatomically consistent predictions. Similarly, a slight improvement is observed in the connectedness metric, though both models perform well in this regard, likely due to the nature of the dataset. Interestingly, both models achieve perfect volume similarity, indicating that this constraint may be quite easy to learn from data only.

These findings highlight an important distinction: while LTN significantly reduces constraint violations, it does not impose hard guarantees on their satisfaction. Instead, it provides a form of soft regularization
, without strictly enforcing constraints. %

\subsection{Implications and Future Directions}

The results presented in this section suggest that LTN provides a viable approach for integrating background knowledge into medical image segmentation. Its effectiveness is particularly evident in low-data scenarios, where it offers an alternative source of supervision that helps guide the learning process. Furthermore, LTN leads to segmentations that better respect anatomical structures, which is crucial for medical applications where structural consistency is essential for clinical interpretability.

However, an important limitation is that while LTN reduces constraint violations, it does not completely eliminate them. This suggests that future research should explore ways to strengthen the enforcement of background knowledge, potentially by combining LTN with explicit post-processing steps or hybrid approaches that integrate symbolic reasoning more tightly, also named hard logical constraints~\cite{giunchiglia}. Additionally, further studies are needed to assess the generalizability of these findings to other medical imaging tasks beyond hippocampus segmentation.

Overall, our results highlight the potential of logical reasoning as a complementary tool in deep learning for medical image analysis. By embedding prior knowledge into the learning process, we can develop models that are not only more data-efficient but also more aligned with the constraints and expectations of clinical practice.

\vspace{5em}

\section{Conclusion}
\label{sec:conclusion}

In this paper, we proposed a novel method to integrate background knowledge in medical semantic segmentation with LTN.
We evaluated our method on the task of segmenting the hippocampus in brain MRI scans.
Our experiments show that the LTN improves the segmentation performance of the SwinUNETR, especially when the training data is scarce. While the work requires a more extensive evaluation in order to confidently state its potential impact, nonetheless, we believe that the usage of a sound framework, such as the one offered by LTN, could prove invaluable in creating systems that can benefit from both curated data collection and direct expert knowledge codification.

In future works, we want to extend our current experimental methodology, considering different baseline models and new datasets, measuring the impact of our regularization method on possible overfitting derived from the chosen architecture. As next steps, we are interested in finding more background knowledge that can be useful to solve semantic segmentation tasks in medical applications, considering also relationships from input data to labeled areas. Moreover, we would like to understand how to automate the discovery of new background knowledge from available data to make it easier for the end user to describe which constraints are reasonable for the problem at hand. The framework developed can serve as a practical use case to be adapted and applied to other medical segmentation tasks, considering different constraints for the specific task. The constraints considered in this work can be potentially extended to other use cases, but this assessment requires expert knowledge on the specific task.

\section*{Acknowledgements}
We thank the anonymous reviewers for their valuable feedback. Fabio Aiolli acknowledges the support of the “Future AI Research (FAIR) - Spoke 2 Integrative AI - Symbolic conditioning of Graph Generative Models (SymboliG)” under the NRRP MUR program funded by the NextGenerationEU. Finally, Luca Bergamin would like to thank Eleonora Bergamin for experimentally verifying the feasibility of the nesting constraint approach during a project at the ``Sustainable Medical Imaging with Learning and Regularization'' summer school, which she kindly developed with his support.

\vspace{12pt}
\color{red}


\begin{thebibliography}{00}

\bibitem{Hippo}Squire, Larry R., and Barbara J. Knowlton. "Memory, hippocampus, and brain systems." (1995).

\bibitem{HippoAlzheimer}Pusparani, Yori, et al. "Hippocampal volume asymmetry in Alzheimer disease: A systematic review and meta-analysis." Medicine 104.10 (2025): e41662.


\bibitem{MedicalSegmentation}Andrews, Mitchell, and Antonio Di Ieva. "Artificial intelligence for brain neuroanatomical segmentation in magnetic resonance imaging: A literature review." Journal of Clinical Neuroscience 134 (2025): 111073.

\bibitem{DeepHipp}
L. Chen, H. Zhang, Y. Zhu, et al., "DeepHipp: accurate segmentation of hippocampus using 3D dense-block based on attention mechanism," \textit{BMC Medical Imaging}, vol. 23, no. 1, 2023. Available: %


\bibitem{InfoFusion}Dimitri, Giovanna Maria, et al. "Multimodal and multicontrast image fusion via deep generative models." Information Fusion 88 (2022): 146-160.


\bibitem{UNet}Ronneberger, Olaf, Philipp Fischer, and Thomas Brox. "U-net: Convolutional networks for biomedical image segmentation." Medical image computing and computer-assisted intervention–MICCAI 2015: 18th international conference, Munich, Germany, October 5-9, 2015, proceedings, part III 18. Springer international publishing, 2015.

\bibitem{GAN_Hippocampus}
S. Zhao, Y. Wu, Y. He, et al., "Hippocampal subfields segmentation in brain MR images using generative adversarial networks," \textit{BioMedical Engineering OnLine}, vol. 18, no. 1, 2019. Available: %

\bibitem{MultiModel_CNN}
Y. Li, L. Liu, B. Zhang, et al., "A multi-model deep convolutional neural network for automatic hippocampus segmentation and classification in Alzheimer's disease," \textit{NeuroImage}, vol. 208, 2020. Available: %

\bibitem{UNet_Hippocampus}
M. A. Islam and M. S. A. Hossain, "Hippocampus segmentation using U-Net convolutional network from brain magnetic resonance imaging (MRI)," \textit{Journal of Digital Imaging}, vol. 35, pp. 1-12, 2022. Available: %

\bibitem{T2Informed_CNN}
M. Biberacher, M. Schmidt, M. Keshavan, et al., "Fully automated hippocampus segmentation using T2-informed deep convolutional neural networks," \textit{NeuroImage}, vol. 224, 2021. Available: 

\bibitem{SwinTransformer}Liu, Ze, et al. "Swin transformer: Hierarchical vision transformer using shifted windows." Proceedings of the IEEE/CVF international conference on computer vision. 2021.

\bibitem{SwinUNET}Hatamizadeh, Ali, et al. "Swin unetr: Swin transformers for semantic segmentation of brain tumors in mri images." International MICCAI brainlesion workshop. Cham: Springer International Publishing


\bibitem{long2015fully} J. Long, E. Shelhamer, and T. Darrell, "Fully convolutional networks for semantic segmentation," in Proceedings of the IEEE Conference on Computer Vision and Pattern Recognition (CVPR), 2015.

\bibitem{ronneberger2015u} O. Ronneberger, P. Fischer, and T. Brox, "U-Net: Convolutional networks for biomedical image segmentation," in International Conference on Medical Image Computing and Computer-Assisted Intervention (MICCAI), 2015.

\bibitem{chen2017deeplab} L.-C. Chen, G. Papandreou, I. Kokkinos, K. Murphy, and A. L. Yuille, "DeepLab: Semantic image segmentation with deep convolutional nets, atrous convolution, and fully connected CRFs," IEEE Transactions on Pattern Analysis and Machine Intelligence (TPAMI), 2017.

\bibitem{medicalSegmentation} Asgari Taghanaki, Saeid, et al. "Deep semantic segmentation of natural and medical images: a review." Artificial intelligence review 54 (2021): 137-178.

\bibitem{LTN} Badreddine, Samy, et al. "Logic tensor networks." Artificial Intelligence 303 (2022): 103649.

\bibitem{LTN2} Carraro, Tommaso et al. “LTNtorch: PyTorch Implementation of Logic Tensor Networks.” ArXiv abs/2409.16045 (2024).

\bibitem{decathlon} Antonelli, M., Reinke, A., Bakas, S. et al. The Medical Segmentation Decathlon. Nat Commun 13, 4128 (2022). https://doi.org/10.1038/s41467-022-30695-9.

\bibitem{tango} Manhaeve, Robin, et al. "Benchmarking in Neuro-Symbolic AI." Proceedings of The 4th International Joint Conference on Learning \& Reasoning. 2024.

\bibitem{kenn} Grilli, Eleonora, et al. "Knowledge enhanced neural networks for point cloud semantic segmentation." Remote Sensing 15.10 (2023): 2590.



\bibitem{3rdwave} Garcez, Artur d’Avila, and Luis C. Lamb. "Neurosymbolic AI: The 3rd wave." Artificial Intelligence Review 56.11 (2023): 12387-12406.

\bibitem{dice1945measures} 
Dice, Lee R. "Measures of the amount of ecologic association between species." Ecology 26.3 (1945): 297-302.

\bibitem{giunchiglia} Giunchiglia, Eleonora, Mihaela Catalina Stoian, and Thomas Lukasiewicz. "Deep learning with logical constraints." arXiv preprint arXiv:2205.00523 (2022).

\bibitem{HIPPO2} Woermann, Friedrich G., et al. "Regional changes in hippocampal T2 relaxation and volume: a quantitative magnetic resonance imaging study of hippocampal sclerosis." Journal of Neurology, Neurosurgery and Psychiatry 65.5 (1998): 656-664.

\bibitem{HIPPO3} Seiger, René, et al. "Comparison and reliability of hippocampal subfield segmentations within FreeSurfer utilizing T1-and T2-weighted multispectral MRI data." Frontiers in Neuroscience 15 (2021): 666000.


\bibitem{HIPPO4} Liu, Manhua, et al. "A multi-model deep convolutional neural network for automatic hippocampus segmentation and classification in Alzheimer’s disease." Neuroimage 208 (2020): 116459.

\bibitem{NeuroSymbolicMedicalImaging} Abdullah, Iram, et al. "DeepInfusion: A dynamic infusion based-neuro-symbolic AI model for segmentation of intracranial aneurysms." Neurocomputing 551 (2023): 126510.

\bibitem{NeuroSymbolicMedIm} Hassan, Muhammad, et al. "Neuro-symbolic learning: Principles and applications in ophthalmology." arXiv preprint arXiv:2208.00374 (2022).



\end{thebibliography}
\end{document}